\pdfoutput=1
\documentclass[11pt]{article}

\usepackage[final]{acl}

\usepackage[T1]{fontenc}    % use 8-bit T1 fonts
\usepackage{hyperref}       % hyperlinks
\usepackage{url}            % simple URL typesetting
\usepackage{booktabs}       % professional-quality tables
\usepackage{amsfonts}       % blackboard math symbols
\usepackage{nicefrac}       % compact symbols for 1/2, etc.
\usepackage{microtype}      % microtypography
\usepackage{xcolor}         % colors
\usepackage{amsthm}
\usepackage{amsmath}
\usepackage{amssymb}
\usepackage{mathrsfs}
\usepackage{caption}
\usepackage{subcaption}
\usepackage{multirow}
\usepackage{color}
\usepackage{tabularx}
\usepackage{arydshln}
\usepackage{graphicx}
\usepackage{bm}
\usepackage{svg}
\usepackage{makecell}
\usepackage{xcolor}  % Add this to your preamble if not already included

\newtheorem{theorem}{Theorem}[section]
\newtheorem{definition}[theorem]{Definition}

\usepackage{hyperref}
\usepackage{url}
\usepackage{amsmath}
\usepackage{amssymb}
\usepackage{amsmath,amsfonts,bm}

\def\eqref#1{equation~\ref{#1}}
\def\1{\bm{1}}

\DeclareMathAlphabet{\mathsfit}{\encodingdefault}{\sfdefault}{m}{sl}
\SetMathAlphabet{\mathsfit}{bold}{\encodingdefault}{\sfdefault}{bx}{n}

\DeclareMathOperator*{\argmax}{arg\,max}

\title{Breach in the Shield: Unveiling the Vulnerabilities of Large Language Models}

\author{
    Runpeng Dai$^{1*}$\quad
    Run Yang$^{2*}$ \quad
    Fan Zhou$^{3}$ \quad
    Hongtu Zhu$^{1\dag}$ \\
    $^{1}$University of North Carolina at Chapel Hill \quad
    $^{2}$BiliBili \\
    $^{3}$Shanghai University of Finance and Economics \\
    \texttt{\{runpeng, htzhu\}@email.unc.edu} \\
    \texttt{yangrun@bilibili.com} \\
    \texttt{zhoufan@mail.shufe.edu.cn}
}

\begin{document}
\maketitle
\begin{abstract}
Large Language Models (LLMs) and Vision-Language Models (VLMs) have achieved impressive performance across a wide range of tasks, yet they remain vulnerable to carefully crafted perturbations. In this study, we seek to pinpoint the sources of this fragility by  identifying parameters and input dimensions (pixels or token embeddings) that are susceptible to such perturbations. 
To this end, we propose a stability measure called \textbf{FI}, \textbf{F}irst order local \textbf{I}nfluence, which is rooted in information geometry and quantifies the sensitivity of individual parameter and input dimensions. Our extensive analysis across LLMs and VLMs (from 1.5B to 13B parameters) reveals that: (I) A small subset of parameters or input dimensions with high FI values disproportionately contribute to model brittleness. (II) Mitigating the influence of these vulnerable parameters during model merging leads to improved performance.
\end{abstract}

\section{Introduction}
\footnote{$^*$Both authors contributed equally to this work.}

Large Language Models (LLMs) and Vision Language Models (VLMs) such as GPT \citep{brown2020language} and Llama \citep{touvron2023llama}, have revolutionized the field of Natural Language Processing (NLP), exhibiting remarkable proficiency across a variety of tasks \citep{gong2024advancing, zheng2025asymmetric, luo2025beyond} and modalities \citep{bai2023qwen, liu2024visual, zheng2025learning}.
These modern LLMs are massive in size, trained on vast amounts of data, and meticulously aligned to prevent from generating harmful content \citep{perez2022red}, leaking private information \citep{zhang2024privacyasst}, or exhibiting sexual or religious bias \citep{xie2023empirical}. 

Despite the enthusiasm for these integrative approaches, a critical issue remains: LLMs remain susceptible to both external and internal perturbations, affecting their reliability and performance.

\begin{figure}[!hbt]
\centering
\includegraphics[width=\columnwidth]{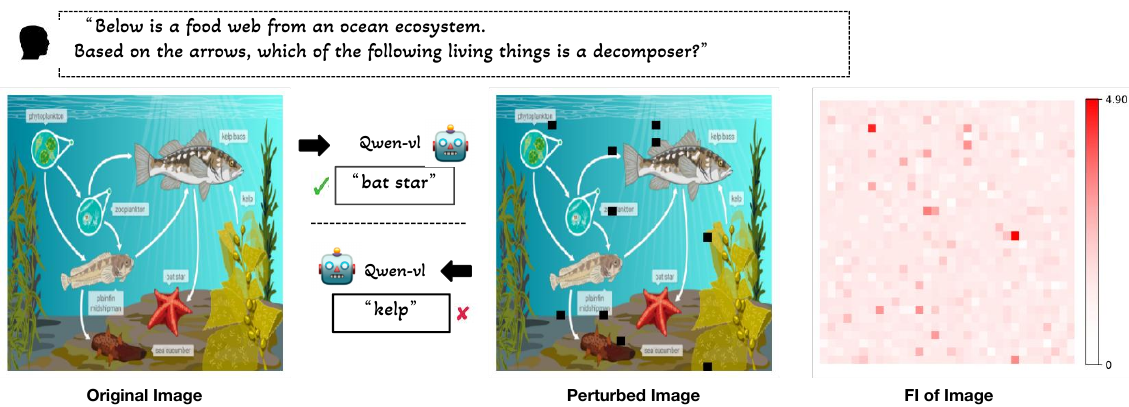}
\caption{A case study of the Qwen-VL model \citep{bai2023qwen} on SCI-QA. The image on the far right visualizes the per-pixel FI values. Masking just 10 pixels with the highest FI values leads to a failure in producing the correct answer.}
\label{vlm-fi}
\end{figure}

\textbf{Externally}, LLMs are vulnerable to input perturbations, such as Embedding-Corrupted Prompts \citep{fort2023scaling, liu2024large}. This susceptibility extends to visual inputs in VLMs, where adversarially optimized images can drastically alter model behavior \citep{qi2024visual}. Beyond adversarial attacks, VLMs exhibit high sensitivity to perturbations in specific local regions of an image—a common issue, as user-uploaded images often suffer from blurring, masking, or low resolution. The vulnerability is highlighted in our case study of the Qwen-VL model. As depicted in Figure \ref{vlm-fi}, masking the ten most sensitive pixels, which are unrelated to the question, resulted in incorrect model outputs.

\begin{figure}[!hbt]
\centering
\includegraphics[width=0.9\columnwidth]{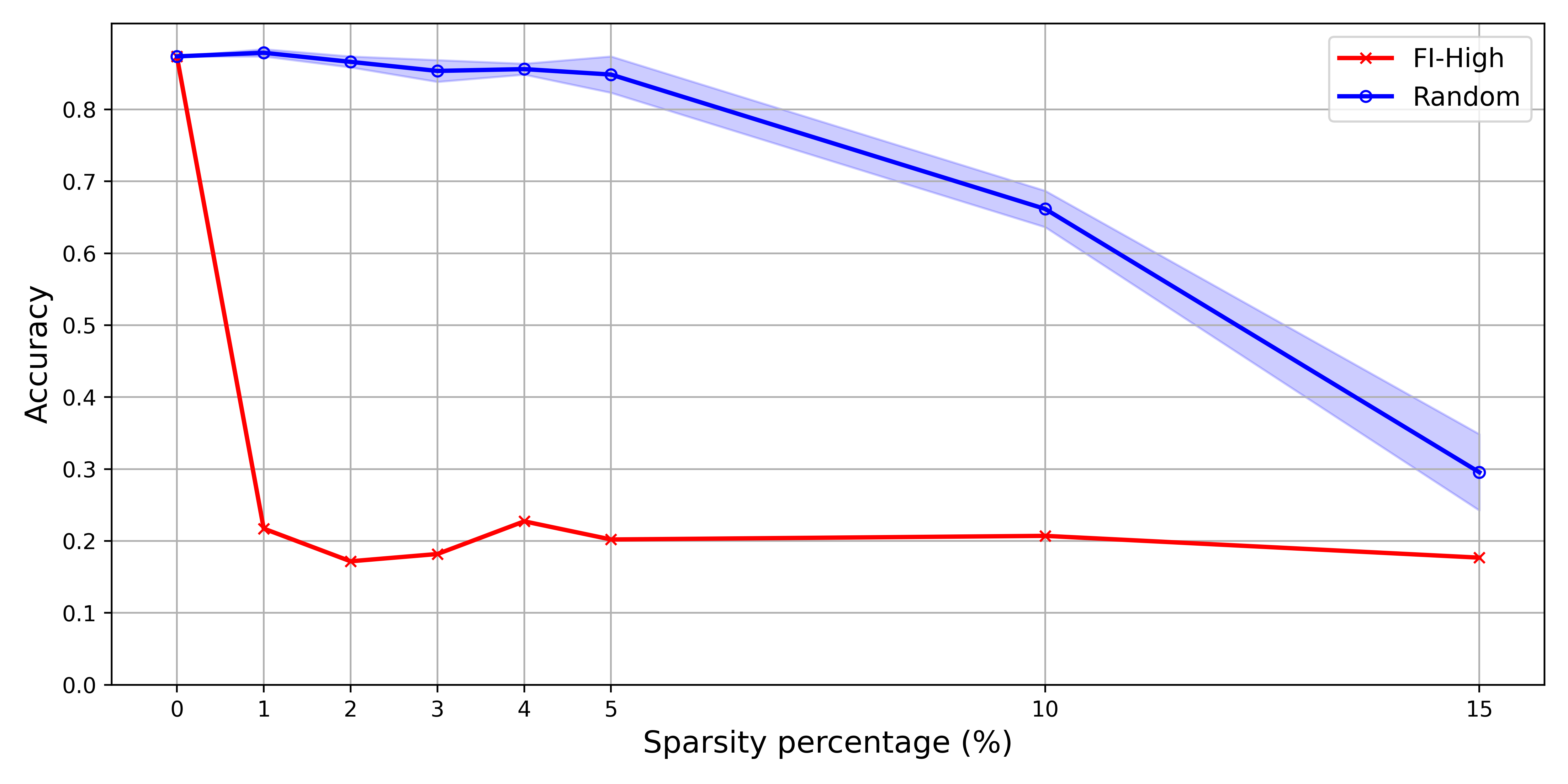}
\caption{A case study of Qwen2 on MMLU-geography. This figure illustrates the decline in response accuracy as a greater percentage of parameters are zeroed. ``FI-High'' targets parameters with the highest FI values, while ``Random'' indicates random parameter removal.}
\label{llm-fi}
\end{figure}

\textbf{Internally}, LLM stability is further challenged by parameter perturbations, often introduced through model merging and quantization. While these techniques improve deployment efficiency by reducing inference costs \citep{frantar2023sparsegpt, ashkboos2024slicegpt}, they can also induce hallucinations and degrade performance \citep{men2024shortgpt, yu2024language, li2024dawn}. However, our findings reveal that parameter susceptibility varies significantly. As Figure \ref{llm-fi} illustrates, randomly dropping 5\% of parameters has a minimal impact on performance. In contrast, zeroing out just 1\% of the parameters identified by our measure can drastically reduce accuracy, even below random guessing levels.

To pinpoint the sources of this fragility, we propose a novel stability measure called \textbf{FI}, \textbf{F}irst order local \textbf{I}nfluence, to quantitatively assess the stability of LLMs against perturbations.
Specifically, we construct a perturbation manifold that encompasses all perturbed models, along with its associated geometric properties.
Our stability measure quantifies the degree of local influence of a perturbation on a given objective function within this manifold, thereby reflecting the stability of individual LLM components.

\textbf{FI's versatility} allows for effective stability assessment under both external and internal perturbations across various granularities—from individual parameters to input features like pixels and patches.

\textbf{FI effectively identifies vulnerabilities}. Our extensive studies validate its effectiveness in pinpointing fragile pixels in VLM vision inputs, vulnerable embedding dimensions of tokens in LLMs (Section \ref{sec: ext}), and salient model parameters (Subsection \ref{sec: dropping}).

% \textbf{FI offers insights for model optimization}. Furthermore, we demonstrate that identifying these vulnerabilities can optimize model performance. Using model merging as a showcase, we show that protecting salient parameters with high FI values significantly mitigates degradation caused by merging (Subsection \ref{sec: merging})

\textbf{FI offers insights into improving model robustness}. We further illustrate that understanding these vulnerabilities can lead to enhanced model resistance to perturbations. By focusing on model merging as an example, we show that safeguarding key parameters identified by high FI values can substantially reduce performance degradation during the merging process (Subsection \ref{sec: merging}).

\section{Related Work}
Recent efforts to evaluate LLM stability typically adopt a coarse-grained approach, aiming to assess the overall robustness of models under various perturbations. One line of work investigates how stability is influenced by sampling parameters, such as temperature, which affect output variability during generation \citep{atil2024llm, ouyang2025empirical}. Another direction studies model sensitivity to input or parameter perturbations. For instance, \citet{liustability} analyze input-level robustness using optimal transport to quantify a model's response to distributional shifts in prompts. On the other hand, \citet{peng2024navigating} focus on parameter-space perturbations, demonstrating that LLMs remain robust to weight changes up to a certain threshold, beyond which performance significantly degrades. They estimate a model’s robustness tolerance by injecting random perturbations into model weights and evaluating the performance drop.

Despite these contributions, fine-grained analyses—such as those examining the effect of individual input tokens, pixels, or specific model parameters—remain underexplored. \citet{wei2024assessing} take a step in this direction by leveraging pruning-based techniques, including SNIP \citep{lee2018snip} and Wanda \citep{sun2023simple}, to identify critical neurons and low-rank structures that impact model safety and utility. However, there is still a lack of unified metrics or frameworks that assess stability with respect to both input- and parameter-level perturbations. Moreover, the downstream applications of such stability assessments remain largely unexamined.

\section{Stability Measure of Large language models}\label{Influence Measure}

In this section, we propose a new metric called FI to quantify the stability of large language models against local perturbations. Considering the auto-regressive nature of LLMs, we first develop FI for single-step generation and discuss its theoretical and computational properties in detail. We then show how FI can be naturally extended to sequence generation tasks. Finally, we compare FI to existing stability measures, highlighting its unique advantages.

\subsection{FI Metric}
\paragraph{\textbf{Problem formulation.}}
Consider an LLM parameterized by $\theta$, with input data $x$, which may consist of text or, for visual language models, a combination of text and images. Given $x$, the model generates a probability distribution over its vocabulary to predict the next token, which can be framed as a classification problem with $K$ classes, where $K$ represents the vocabulary size.

However, vocabulary sizes are typically large \citep{bai2023qwen, dubey2024llama}, and predictions are often concentrated on a small subset of tokens. Instead of using the entire vocabulary, it is more efficient to focus on a relevant subset based on the task. For example, in multiple-choice questions, probabilities are restricted to the choices "A", "B", "C", or "D".  Classes can also be defined semantically, such as categorizing tokens as "neutral" or "notorious" in toxicity detection \citep{gehman2020realtoxicityprompts}. 

With appropriately defined classes, the predicted probability for class $y \in \{1, \ldots, K\}$ is denoted as $P(y|x,\theta)$, satisfying $\sum_{y=1}^K P(y|x,\theta) = 1$. Let $\omega\in \mathbb{R}^d$ be a perturbation vector varies in an open subset $\Omega$. $\omega$ can be applied to a subset of the model parameters $\theta$ and locations within the input data $x$. We denote the output of the perturbed model under this perturbation as $P(y|x, \theta, \omega)$.

\paragraph{\textbf{Perturbation Manifold and FI}}Since our primary interest lies in examining the behavior of $P(y|x, \omega, \theta)$ as a function of $\omega$ near $\omega_0 = 0$, we shift focus from $\theta$ to $\omega$. We introduce the perturbation manifold as defined in \cite{zhu2007perturbation} and \cite{zhu2011bayesian}. 

 \begin{definition}
    Define the $d$-dimensional perturbation manifold $\mathcal{M} =  \{ P(y | x,  \theta, \omega) : \omega \in \Omega \}$, which encompasses all perturbed models. Assume that for all $\omega \in \Omega$, the perturbed models $\{P(y = i | x,  \theta, \omega)\}_{i=1}^K$ are positive and sufficiently smooth. The tangent space $T_{\omega}$ of $\mathcal{M}$ at $\omega$ is spanned by the partial derivatives of the log-likelihood function $\ell (\omega | y,  x,  \theta) = \log  P(y | x,  \theta,  \omega)$ with respect to $\omega$, specifically $T_{\omega} = \text{span}\{ \frac{\partial }{\partial \omega_{i}} \ell (\omega | y,  x,  \theta) \}_{i=1}^{d} $.
\end{definition}

The metric \( g_{\omega} \) on \( \mathcal{M} \) can be defined with the metric tensor \( G_{\omega} \).
Consider two tangent vectors at \( \omega \) given by $v_j(\omega) = h_j^\top \partial_\omega \ell (\omega | y,  x,  \theta) \in T_{\omega}$, where $h_1$ $h_2$ are the weights on the basis. Their inner product is defined as:
\[
 \langle v_1(\omega), v_2(\omega) \rangle_{g_{\omega}} =
 \sum_{y=1}^{K} v_1(\omega) v_2(\omega) P(y | x,  \theta).
\]
The metric tensor \( G_{\omega} \) is given by:
\[
 G_{\omega} = \sum_{y=1}^{K}  \partial_\omega \ell (\omega | y,  x,  \theta) \partial^\top_\omega \ell (\omega | y,  x,  \theta) P(y | x,  \theta, \omega).
\]
Subsequently, the norm of \( v_j(\omega) \) under metric \( g_{\omega} \) is $\|v_j\|_{g_{\omega}} = \sqrt{h_j^\top G_{\omega} h_j}$.
Let \( C(t) = P(y|x, \theta, \omega(t)) \) be a smooth curve on the manifold \( \mathcal{M} \) connecting two points
\( \omega_1 = \omega(t_1) \) and \( \omega_2 = \omega(t_2) \). Then, the distance between \( \omega_1 \) and \( \omega_2 \) along the curve \( C(t) \) is given by:
\begin{equation*}
\begin{aligned}
S_{C} \left({\omega}_{1}, {\omega}_{2}\right)
&= \int_{t_{1}}^{t_{2}} \sqrt{\|\partial_t \log P(y|x, \theta, \omega(t) )\|_{g_{\omega}}} \, d t \\
&= \int_{t_{1}}^{t_{2}} \sqrt{\frac{d {\omega}(t)^{T}}{d t} G_{\omega(t)} \frac{d {\omega}(t)}{d t}} \, d t.
\end{aligned}
\end{equation*}

With the Perturbation manifold $\mathcal{M}$ and respective metric $g_{\omega}$ defined, we are ready to propose the metric that quantifies the stability of large language models (LLMs) against various types of local perturbations.
Let $f(\omega)$ be the objective function of interest for sensitivity analysis, in our case being $-\log P(y_{pred}|x, \theta, \omega)$, we can define the following (first-order) local influence metric FI:
%which quantifies the robustness of large language models (LLMs) against various types of local influences.
\begin{definition}\label{def2.2}
Given the perturbation manifold $\mathcal{M}$ and its metric, the first-order local stability measure of $f({\omega})$ at $\omega(0)=\omega_0$ is defined as
\begin{equation}\label{fi1}
\mathbf{FI}_{{\omega}}\left(\omega_0\right)=\max _{C} \lim _{t \rightarrow 0} \frac{[f({\omega}(t))-f({\omega}(0))]^{2}}{S_{C}^{2}({\omega}(t), {\omega}(0))}.    
\end{equation}
\end{definition}

The ratio in Equation \ref{fi1} measures the amount of change introduced to the objective function relative to the distance of the perturbation on the perturbation manifold. Thus, Equation \ref{fi1} can be naturally interpreted as the maximum local ratio of change among all possible perturbation curves \( C(t) \). 

\paragraph{\textbf{Computation of FI.}} As we will show, Theorem \ref{invariant} on diffeomorphic reparameterization invariance enables us to derive an easy-to-compute solution for Equation \ref{fi1}, while addressing the low-dimensionality problem inherent in LLMs.

\begin{theorem} \label{FI-close}
If \( G_{\omega} \) is positive definite, the \textbf{FI} measure has the following closed-form:
\begin{equation}
\mathbf{F} \mathbf{I}_{{\omega}}\left({\omega}_{0}\right) =
{\nabla}_{f\left({\omega}_{0}\right)}^{T} G^{-1}_{\omega_0} {\nabla}_{f\left({\omega}_{0}\right)},
\label{fi-final}
\end{equation}
where 
\[
{\nabla}_{f\left({\omega}_{0}\right)}=\frac{\partial f({\omega})}{\partial {\omega}} \Big|_{{\omega}={\omega}_{0}}.
\]
\end{theorem}

The detailed proof of Theorem \ref{FI-close} can be found in Appendix \ref{proof2.4}. It is important to note that the closed form of FI in Theorem \ref{FI-close} depends on the positive definiteness of \( G_{\omega} \), which is not always guaranteed. This is due to the fact that the parameters in LLMs are often high-dimensional tensors with low-rank structures \citep{kaushal2023lord}.

We apply the invariant Theorem \ref{invariant} by transforming \( {\omega} \) to a vector \( {\nu} \) such that \( G_{\nu}= \mathbf{I}_K \), where \( K \) is an integer. Specifically, we notice that $G_{\omega_{0}} = {B}_{0}^{T}{B}_{0}$, where
\[
{B}_{0} = \big[ P(y=i | x,  \theta, \omega)^{1/2} \partial_\omega \ell (\omega | y =i,  x,  \theta) \big]_{i \leqslant K }.
\]
Let \( r_0 = \text{rank}(G_{\omega_0}) \), we apply the compact SVD to \( {B}_{0} \in \mathbb{R}^{p \times K} \), which yields ${B}_{0} = {V}_{0}\Lambda_{0}{U}_{0}$, where \( V_{0} \in \mathbb{R}^{p\times r_0} \) and \( U_{0} \in \mathbb{R}^{r_0\times K} \) are semi-orthogonal matrices and \( \Lambda_{0} \in \mathbb{R}^{r_0\times r_0} \) is a diagonal matrix. Under the transformation \( \nu = {\Lambda}_{0}{V}_{0}^{T}{\omega} \), we have $\mathbf{FI}_{{\omega}}\left({\omega}_{0}\right) = \mathbf{FI}_{{\nu}}\left({\nu}_{0}\right)$ , which can be expressed as
\begin{align*}
{\nabla}_{f\left({\omega}_{0}\right)}^{\top} 
   \left( {V}_{0} {R}_{0} \right)^{\top} 
   {\Lambda}_{0}^{-2} 
   \left( {V}_{0} {R}_{0} \right) 
   {\nabla}_{f\left({\omega}_{0}\right)},
\end{align*}
where the equality holds by applying the chain rule to \( G_{\nu} \).

\paragraph{\textbf{FI for sequence generation.}}
Sequence generation is essentially multiple rounds of next-token generation, where the $l$-th token $y^{(l)}$ is generated given the initial input $z$ and previously generated tokens $\bm y^{(l)} = \{y^{(1)}, \ldots, y^{(l-1)}\}$. We define the FI measure for generating the $l$-th token $y^{(l)}$ given the initial input $z$ by averaging out the randomness from the preceding steps $\mathbf{FI}_l(z) = \mathbb{E}_{\bm y^{(l)}}[\mathbf{FI}(\{z, \bm y^{(l)}\}, \theta, \omega) | z]$.

To formulate an overall measure for sequence generation, we aggregate these per-token FI measures. Since sequences generated by LLMs can vary in length, we propose two methods to handle this heterogeneity. The first approach sets a fixed horizon $L$ and computes the mean FI over these rounds
\begin{equation} \mathbf{FI}^L_{\text{seq}}(z) = \frac{1}{L} \sum_{l=1}^L\mathbf{FI}_l(z). \label{FI-seq}
\end{equation} 
Alternatively, inspired by the concept of average discounted rewards in reinforcement learning \citep{liu2018breaking}, we consider sequences of potentially infinite length and propose a discounted FI measure with discount factor $\gamma$ 
\begin{equation*} \mathbf{FI}^{\infty, \gamma}_{\text{seq}}(z) = (1-\gamma)\sum_{l=0}^{\infty} \gamma^{l} \cdot \mathbf{FI}_l(z). \end{equation*}
By taking the expectation over the distribution of $z$, we obtain the average FI for sequence generation in both cases $\mathbb{E}_{P_z}[\mathbf{FI}^L_{\text{seq}}(z)]$ and $\mathbb{E}_{P_z}[ \mathbf{FI}^{\infty, \gamma}_{\text{seq}}(z) ]$, respectively.

\subsection{Other Measures \& Discussion}
We note that several alternative methods can also serve as stability measures for LLMs. We provide their explicit formulations and compare them with FI.

\noindent\textbf{Jacobian Norm \citep{novak2018sensitivity}:}
\vspace{-1.2ex}
\begin{equation*}
    \|\partial_{\omega} f(y_{pred},\omega)\|_2
\end{equation*}

\noindent\textbf{SNIP \citep{lee2018snip}:}
\vspace{-1.2ex}
\begin{equation*}
     \|\omega \odot \partial_{\omega} f(y_{pred},\omega)\|_2
\end{equation*}

Both measures focuses solely on $y_{pred}$, while neglecting the probabilities assigned to other choices. For example, consider two output distributions: (0.9, 0.05, 0.05, 0.02) and (0.3, 0.25, 0.25, 0.2). In both cases, the model selects option A. However, the second distribution is more unstable, as a small perturbation in the probabilities could lead to a different prediction. In contrast, FI measure accounts for both the probability and gradient across all possible choices.

\noindent\textbf{Saliency map \citep{simonyan2013deep}:} 
\vspace{-1.2ex}
\begin{equation*}
\begin{cases}
0 ~~ \text{if} ~\frac{\partial f(y_{pred},\omega)}{\partial \omega} < 0 \text{ or } \sum_{y \neq y_{pred}} \frac{\partial f(y,\omega)}{\partial \omega} > 0 \\
- \frac{\partial f(y_{pred},\omega)}{\partial \omega} \sum_{y \neq y_{pred}} \frac{\partial f(y,\omega)}{\partial \omega}
\end{cases}
\end{equation*}

Saliency map take the gradient of all choices into account. But loose too much information by zeroing out many of the gradient. 

Although all these methods can be used to assess the vulnerability of LLMs, we highlight FI for its distinct advantages. \textbf{Effectiveness:} A quantitative comparison of these measures is provided in Section~\ref{sec: ext}, while their computational complexities are discussed in Appendix~\ref{Appendix: computation}. \textbf{Theoretical rigor:} In particular, only FI possesses a reparameterization invariance property (see Appendix~\ref{appendix: invariance}), which further distinguishes it by enhancing interpretability.

\section{External perturbations analysis}\label{epa}
\label{sec: ext}
In this section, we first demonstrate the effectiveness of FI in identifying vulnerable locations in both vision and language inputs though guided attack. Then, we conclude the section with a finding from cross-modal analysis.

\textbf{Identify Fragile Pixels} We conduct the attack process on the MMbench dataset \cite{liu2024mmbench}, a comprehensive benchmark designed to evaluate various multi-modal capabilities of VLMs. For a fair comparison, we identify the top 10 pixels using different stability measures and assess the model's performance after masking out the corresponding pixels.

% {\color{blue}From Table~\ref{vlm-mmbench}, we observe that, masking out pixels identified by the FI measure leads to the most significant performance degradation, clearly demonstrating the superiority of FI over other measures.}
\begin{table}[!hbt]
\caption{Accuracy on the MMBench dataset after masking out top ten pixels in images identified by different measures. The first block shows summary accuracy across all models and methods.}
\label{vlm-mmbench}
\renewcommand\arraystretch{1.2}
\resizebox{\columnwidth}{!}{
\begin{tabular}{c|l|c|c|c|c}
\Xhline{1.2pt}
\textbf{Model} & \textbf{Method} & \makecell[c]{Action\\Recognition} & \makecell[c]{Attribute\\Recognition} & \makecell[c]{Celebrity\\Recognition} & \makecell[c]{Function\\Reasoning} \\
\hline\hline

\multirow{4}{*}{\makecell{Qwen\\VL}} 
    & FI (Ours)         & \textbf{0.320} & \textbf{0.402} & \textbf{0.673} & \textbf{0.411} \\ 
    & Jacobian          & 0.668 & 0.587 & 0.906 & 0.604 \\ 
    & Saliency          & 0.782 & 0.525 & 0.873 & 0.639 \\
    & Random          & 0.812 & 0.550 & 0.881 & 0.683 \\
    & Original          & 0.814 & 0.549 & 0.882 & 0.686 \\
\hline

\multirow{4}{*}{\makecell{Qwen2.5\\VL-3B}} 
    & FI (Ours)         & \textbf{0.720} & \textbf{0.735} & \textbf{0.780} & \textbf{0.723} \\
    & Jacobian          & 0.731          & 0.752          & 0.797          & 0.755 \\
    & Saliency          & 0.745          & 0.761          & 0.797          & 0.774 \\
    & Random          & 0.882          & 0.931          & 0.957          & 0.928 \\
    & Original          & 0.890          & 0.946          & 0.959          & 0.930 \\
\hline

\multirow{4}{*}{\makecell{Qwen2.5\\VL-7B}} 
    & FI (Ours)         & \textbf{0.768} & \textbf{0.750} & \textbf{0.796} & \textbf{0.723} \\
    & Jacobian          & 0.778          & 0.768          & 0.815          & 0.755 \\
    & Saliency          & 0.792          & 0.777          & 0.815          & 0.774 \\
    & Random          & 0.891          & 0.944          & 0.951          & 0.925 \\
    & Original          & 0.890          & 0.946          & 0.959          & 0.930 \\

\Xhline{1.2pt}
\end{tabular}
}
\end{table}

\textbf{Identify Vulnerable Embedding Dimensions} We conduct attack on pure-text LLMs to verify the effectiveness of our approach in identifying vulnerable embedding dimensions. Specifically, we follow the token embedding attack methods proposed in \cite{liu2024large} and \cite{fort2023scaling}.

More concretely, we compute the stability measure for each embedding dimension and select the top 0.1\% most sensitive dimensions($\omega$) as identified by the metrics. We then apply a gradient-based attack strategy following \cite{fort2023scaling}, perturbing the selected dimensions in the direction of $- \nabla_{\omega} \log P(y_{\text{pred}} \mid x, \theta)$.

% {\color{blue}As shown in Table ~\ref{llm-mmlu}, demonstrate the effectiveness of the FI method in identifying and attacking the most sensitive dimensions. The results indicate that FI consistently outperforms other measures in detecting instability. }

From both Table~\ref{vlm-mmbench} and Table~\ref{llm-mmlu}, we observe the following: (I) Stability measures are effective in identifying vulnerable input dimensions (i.e., pixels in images and dimensions in embeddings). Notably, LLMs are generally robust to random perturbations and such perturbations rarely lead to significant performance degradation. In contrast, perturbations guided by stability measures consistently result in substantial drops in performance. (II) Among all the stability measures evaluated, FI proves to be the most effective: masking pixels or perturbing dimensions identified by FI leads to the largest observed decline in performance.

\begin{table}[!hbt]
\caption{Comparison of accuracy in the MMLU dataset after perturbing the same number of dimensions in the embedding space identified using different measures.}
\label{llm-mmlu}
\renewcommand\arraystretch{1.2}
\resizebox{\columnwidth}{!}{
\begin{tabular}{c|l|l|l|l|l}
\Xhline{1.2pt}
\multicolumn{1}{l|}{\textbf{Model}} & \textbf{Method} & \textbf{Business} & \textbf{Geo} & \textbf{Culture} & \textbf{Law} \\ \hline\hline
\multirow{5}{*}{\makecell{Pythia\\1B}}          
& Saliency        & 0.278             & 0.272             & 0.210            & 0.243         \\
& Jacobian       & 0.273             & 0.264             & 0.201            & 0.241         \\
& Random                & 0.301             & 0.368            & 0.237            & 0.246         \\
& FI (ours)           & \textbf{0.270}    & \textbf{0.261}    & \textbf{0.195}   & \textbf{0.236} \\
& SNIP                & 0.297             & 0.281            & 0.226            & 0.242         \\
& Original            & 0.303             & 0.370             & 0.240            & 0.247         \\ \hline

\multirow{5}{*}{\makecell{Qwen2.5\\3B}}         
& Saliency        & 0.677             & 0.637             & 0.632            & 0.560         \\
& Jacobian       & 0.665             & 0.641             & 0.625            & 0.560         \\
& Random          & 0.805             & 0.781             & 0.781            & 0.672         \\
& FI (ours)           & \textbf{0.656}    & \textbf{0.620}    & \textbf{0.610}   & \textbf{0.547} \\
& SNIP                & 0.783             & 0.663             & 0.665            & 0.563         \\
& Original            & 0.810             & 0.800             & 0.785            & 0.673         \\ \hline

\multirow{5}{*}{\makecell{Qwen2.5\\7B}}         
& Saliency        & 0.756             & 0.789             & 0.709            & 0.725         \\
& Jacobian       & 0.764             & 0.782             & 0.717            & 0.720         \\
& Random                & 0.852             & 0.884             & 0.802            & 0.735         \\
& FI (ours)           & \textbf{0.748}    & \textbf{0.780}    & \textbf{0.705}   & \textbf{0.713} \\
& SNIP                & 0.757             & 0.791             & 0.710            & 0.727         \\
& Original            & 0.856             & 0.890             & 0.810            & 0.737         \\
\Xhline{1.2pt}
\end{tabular}
}
\end{table}

\begin{figure*}[!hbt]
\centering
\includegraphics[width=\textwidth]{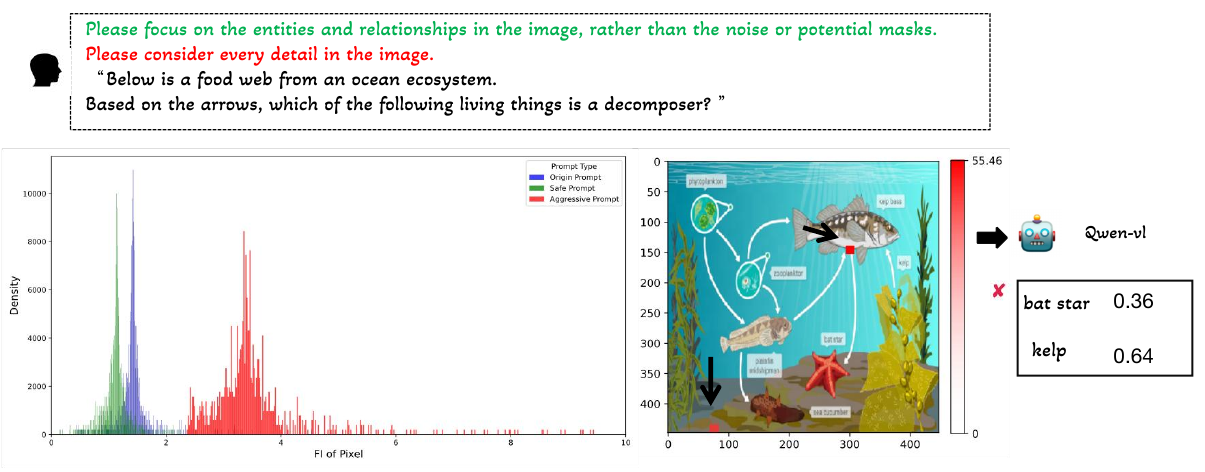}
\caption{A case study utilizing FI for cross-modal analysis. In the same example, the bottom-left image shows how Aggresive and Safe prompts affects the FI distribution on the image. }
\label{vlm-fi-prompt}
\end{figure*}

\textbf{Effect of Prompting on Pixel Vulnerability} 
While the significant impact of prompt design on VLM performance is well-recognized \citep{zhou2022learning}, and carefully crafted prompts are known to even jailbreak these models \citep{shayegani2023jailbreak}, a quantitative analysis of this cross-modal influence – specifically, how prompting affects the processing and stability of visual input – remains largely unexplored.

Our study aims to bridge this gap by investigating how varying prompt instructions influence the sensitivity of VLMs to visual perturbations. Specifically, we examine two types of prompts:
\begin{itemize}
    \item Aggressive Prompts: Designed to encourage the model to consider every detail in the image, potentially increasing sensitivity to noise.
    \item Safe Prompts: Intended to focus the model on salient entities and relationships, potentially enhancing robustness by ignoring irrelevant details.
\end{itemize}

We computed the FI value for each pixel and visualized the resulting distributions under different prompt settings, as illustrated in Figure \ref{vlm-fi-prompt}. Our main findings are as follows: 

\textbf{(I) Prompt choice has a substantial impact on the stability of individual pixels within the image.} As shown on the left of Figure \ref{vlm-fi-prompt}, aggressive prompts shift the FI distribution toward higher values, resulting in a marked increase in both the mean and maximum FI values. This suggests that the model becomes more sensitive to pixel-level perturbations throughout the image. In contrast, safe prompts significantly shift the FI distribution toward lower values, indicating reduced sensitivity and improved stability against perturbations in less relevant regions.

\textbf{(II) Vulnerability remains even with careful prompt design.}
Although safe prompts generally reduce FI values, they do not fully guarantee model stability, as outliers with large FI values persist. As shown in the right column of Figure \ref{vlm-fi-prompt}, even when applying the safe prompt, masking out the two pixels with the highest FI values still leads to incorrect model predictions. This result underscores the persistent challenge of achieving robustness in VLMs and demonstrates the effectiveness of the FI measure for identifying vulnerable regions.

Our findings contribute to the growing body of literature on cross-modal interactions in VLMs, offering a stability-centric perspective that complements existing behavioral and attributional analyses. Importantly, this framework can inform the development of more robust multimodal systems and prompt design strategies for safety-critical applications.

\section{Internal perturbations analysis}\label{internal}

In this section, we first conduct an parameter sparsification experiment to demonstrate the effectiveness of the FI. We then apply the FI measure to mitigate parameter interference during model merging, showcasing its potential for guiding LLM improvement.

\subsection{Parameter sparsification}
\label{sec: dropping}
\begin{figure*}[!hbt]
     \centering
     \begin{subfigure}[b]{.45\textwidth}
         \centering
      \includegraphics[width=0.9\textwidth]{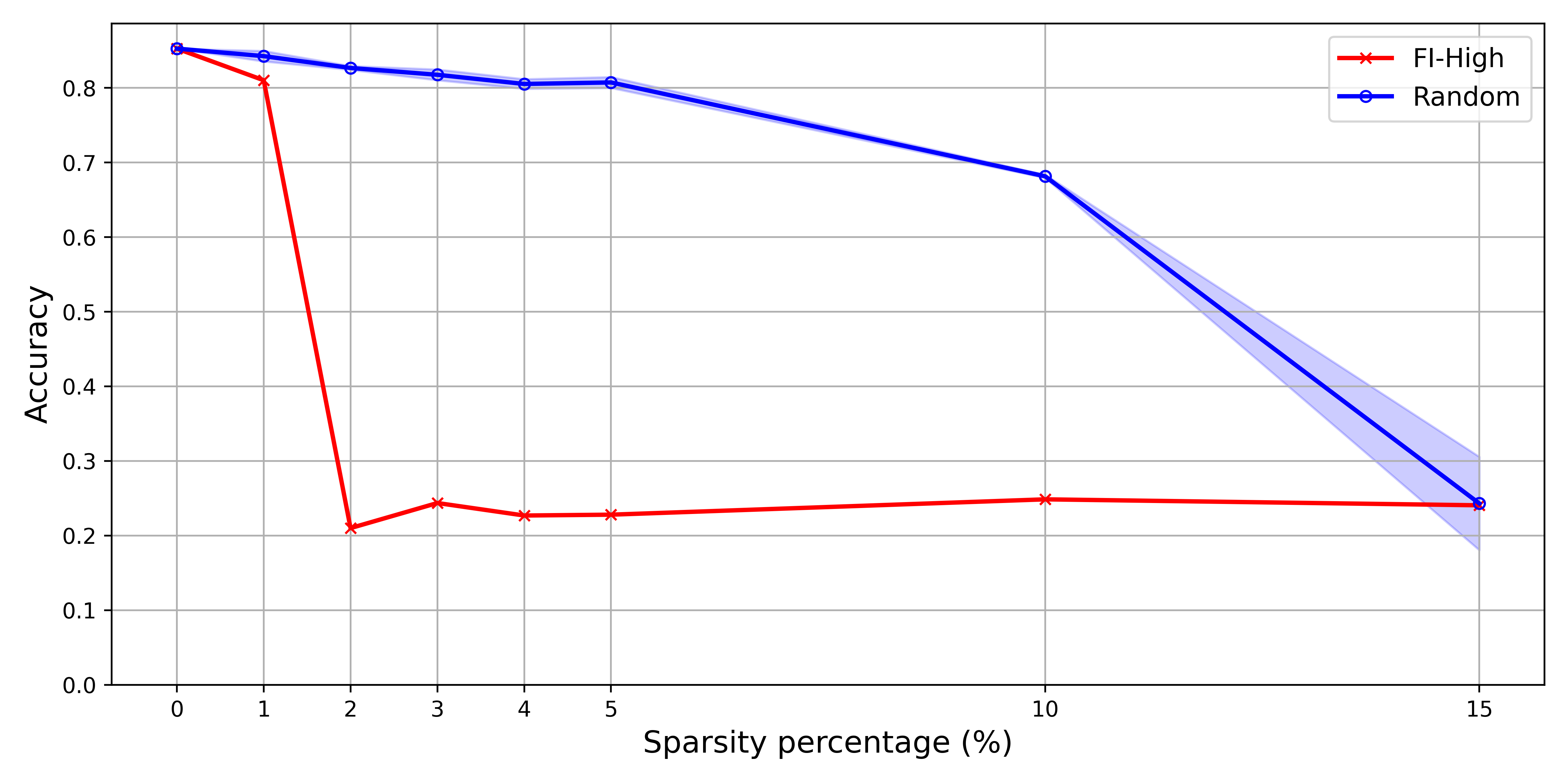}
         \caption{business}
         \label{drop-mmlu-business-1bit}
     \end{subfigure}
     \begin{subfigure}[b]{.45\textwidth}
         \centering
         \includegraphics[width=0.9\textwidth]{fig/drop_geography.png}
         \caption{geography}
         \label{drop-MMLU-Biology}
     \end{subfigure}
     \quad
     \begin{subfigure}[b]{.45\textwidth}
         \centering
         \includegraphics[width=0.9\textwidth]{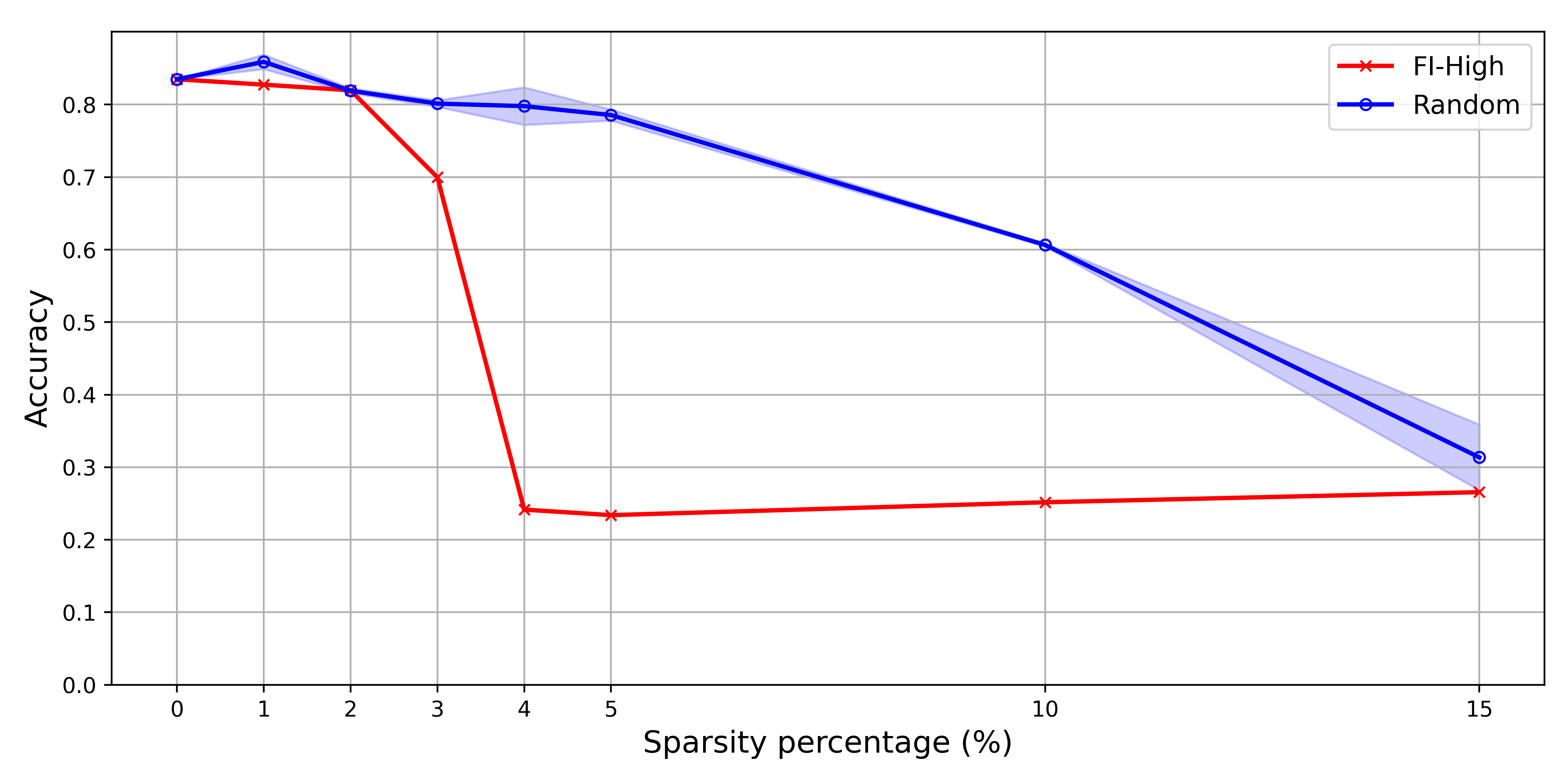}
         \caption{culture}
         \label{drop-mmlu-culture-1bit}
     \end{subfigure}
     \begin{subfigure}[b]{.45\textwidth}
         \centering
         \includegraphics[width=0.9\textwidth]{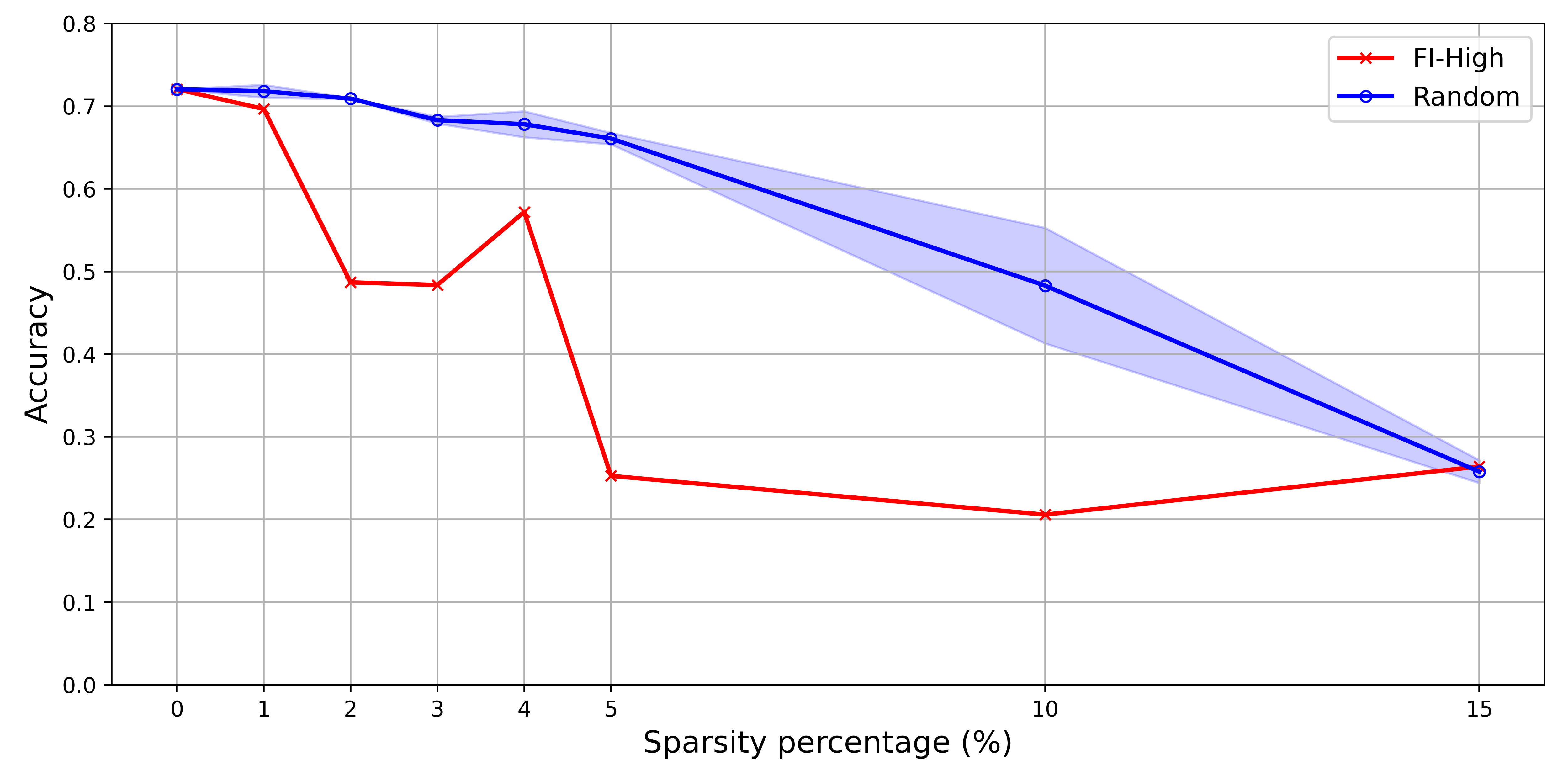}
         \caption{law}
         \label{drop-mmlu-law-4bit}
     \end{subfigure}
     \caption{Comparing the accuracy in the MMLU dataset of Qwen2-7B when parameters have been sparsified at different rates. }
     \label{sparsity-knowledge}
\end{figure*}
We conduct experiments on multiple-choice problems from the MMLU dataset \citep{hendrycks2020measuring} and sequence generation tasks from Alpaca-Eval \citep{dubois2024length} to examine how these perturbations impact two key capabilities of large models: knowledge retention and instruction-following. Details of both experimental setups are provided in Figure~\ref{Appendix: sparsification}.

As shown in Figure~\ref{llm-fi} and Figure \ref{sparsity-knowledge} , sparsifying just 2–3\% of the high-FI parameters significantly degrades the model’s knowledge capacity, leading to catastrophic forgetting and hallucinations, with performance dropping by up to 75\%. A similar trend is observed in Table~\ref{sparsity-seq} at around the 10\% sparsity level. In contrast, models remain relatively robust against random sparsification, often exhibiting nearly identical behavior even after 5\% sparsification.

These findings demonstrate FI's effectiveness in identifying fragile parameters and further support the inherent structure within the parameter matrix, aligning with recent observations on model brittleness \citep{ma2023llm, wei2024assessing, yu2024language}.

\subsection{FI-Guided Parameter Protection in Model Merging}
\label{sec: merging}

Model merging is a technique for acquiring domain-specific knowledge by combining models from different domains, thereby reducing the computational cost of additional fine-tuning (see \citep{yang2024model} for a review). However, a persistent challenge is that merging parameters introduces perturbations that can hinder a model's ability to retain previously learned information. To address this, we use FI to identify parameters susceptible to forgetting and exclude them from the merging process.

We demonstrate that FI can be seamlessly integrated into mainstream model merging methods, including \textbf{Average Merging} \citep{wortsman2022model}, \textbf{Task Arithmetic} \citep{ilharco2022editing}, and \textbf{TIES} \citep{yadav2024ties}. Additionally, we include \textbf{DARE} \citep{yu2024language} as a competing baseline for completeness.

We consider merging two models, $A$ and $B$, both fine-tuned from the same base model. Let $\theta_A$, $\theta_B$, and $\theta_{\text{Base}}$ denote the parameters of models $A$, $B$, and the base model, respectively. We first introduce the merging methods and then demonstrate how FI can be integrated into these algorithms to mitigate perturbation effects.

\textbf{Average Merging}  
Average merging obtains the merged model by averaging $\theta_A$ and $\theta_B$, resulting in parameters  
$\theta_{\text{Avg}} = \frac{\theta_A+\theta_B}{2}$.

\textbf{Task Arithmetic}  
Task arithmetic constructs ``task vectors'' by subtracting a base model from each task-specific model and then merges these vectors linearly before adding back the base model $\theta_{\text{Task}} =\theta_{\text{Base}} + \gamma (\delta_A + \delta_B)$, where \( \delta_A = \theta_A - \theta_{\text{Base}} \) and similarly for \( \delta_B \). 

Both Average Merging and Task Arithmetic modify all parameters in models $A$ and $B$, potentially degrading performance by disturbing their most sensitive parameters. To address this, we employ a protection strategy that preserves these vulnerable parameters while merging only the less critical ones. Specifically, we identify the top $k\%$ of high-FI parameters in both models and record their locations in $\Theta_A$ and $\Theta_B$. Then, for each layer in both $\theta_{\text{Task}}$ and $\theta_{\text{Avg}}$, we revert parameters at locations in $\Theta_A \cap \Theta_B^\complement$ to their original values from $\theta_A$, and parameters at locations in $\Theta_B \cap \Theta_A^\complement$ to their original values from $\theta_B$.

\textbf{TIES} (\textbf{T}r\textbf{I}m, \textbf{E}lect \textbf{S}ign) operates in two steps. First, it reduces redundancy by setting a fraction of the ``task vectors'' $\delta_A$ and $\delta_B$ to zero. Then, for each remaining entry, it retains the weight from the vector with the larger absolute value.

FI-guided protection can be incorporated into both steps. In the first step, we protect $\delta_A$ at locations $\Theta_A$ and $\delta_B$ at $\Theta_B$ from being trimmed. In the second step, entries within $\Theta_A$ are preserved as $\delta_A$, while those in $\Theta_B$ remain as $\delta_B$, regardless of their absolute values. 

We merged \textbf{Qwen2.5-Math-7B} \citep{yang2024qwen2} and \textbf{HuatuoGPT-o1-7B} \citep{chen2024huatuogpt}, as both models are further fine-tuned from the same base model, \textbf{Qwen2.5-7B} \citep{yang2024qwen2math}. We evaluate the performance of the merged models on math and health subjects within the MMLU benchmark \citep{hendrycks2020measuring}.

From Table~\ref{tab:merge}, we observe the following: (1) Across all merging methods, FI-guided protection generally enhances the performance of the merged models in both domains. For example, the Average model merging method with FI-guided protection yields approximately a 1\% improvement in both the Math and Health domains. (2) Furthermore, TIES with FI protection applied in its first stage performs the best among all merging methods, whereas DARE does not perform well in this setting.

\begin{table}[h]
    \centering
    \renewcommand{\arraystretch}{1.2}
    \resizebox{\columnwidth}{!}{
    \begin{tabular}{c|c|c c c}
        \Xhline{1.2pt}
          & \textbf{FI-protect} & \textbf{Math} & \textbf{Health} & \textbf{Mean} \\
        \hline
        \textbf{\makecell{Qwen2.5\\Math-7B}} & / & 0.616 & / & / \\
        \hline
        \textbf{\makecell{Huatuo\\o1-7B}} & / & / & 0.724 & / \\
        \hline
        \multirow{2}{*}{\textbf{Average}} & Without & 0.534 \textcolor{blue}{(-8.2\%)} & 0.514 \textcolor{blue}{(-21.0\%)} & 0.524 \\
        & With & 0.543 \textcolor{cyan}{(-7.3\%)} & 0.522 \textcolor{cyan}{(-20.2\%)} & 0.533 \\
        \hline
        \multirow{2}{*}{\textbf{Task}} & Without & \underline{0.577} \textcolor{blue}{(-3.9\%)} & 0.597 \textcolor{blue}{(-12.7\%)} & \underline{0.587} \\
        & With & 0.573 \textcolor{cyan}{(-4.3\%)} & \underline{0.598} \textcolor{cyan}{(-12.6\%)} & 0.586 \\
        \hline
        \multirow{3}{*}{\textbf{TIES}} & Without & 0.565 \textcolor{blue}{(-5.1\%)} & 0.596 \textcolor{blue}{(-12.8\%)} & 0.581 \\
        & With I & \textbf{0.583} \textcolor{cyan}{(-3.3\%)} & \textbf{0.606} \textcolor{cyan}{(-11.8\%)} & \textbf{0.595} \\
        & With II & 0.566 \textcolor{cyan}{(-5.0\%)} & 0.601 \textcolor{cyan}{(-12.3\%)} & 0.584 \\
        \hline
        \textbf{\makecell{DARE\\Task}} & / & 0.573 \textcolor{blue}{(-4.3\%)} & 0.589 \textcolor{blue}{(-13.5\%)} & 0.581 \\
        \textbf{\makecell{DARE\\TIES}} & / & 0.560 \textcolor{blue}{(-5.6\%)} & 0.588 \textcolor{blue}{(-13.6\%)} & 0.574 \\
        \Xhline{1.2pt}
    \end{tabular}
    }
    \caption{Performance of merging Qwen2.5-Math-7B and HuatuoGPT-o1-7B. The ``Mean'' column reports the average accuracy across both Math and Health tasks. Blue and cyan percentages indicate the performance drop for the ``Without'' and ``With'' variants comparing to the original model, respectively.}
    \label{tab:merge}
\end{table}

\begin{figure}[!hbt]
\centering
\includegraphics[width=\columnwidth]{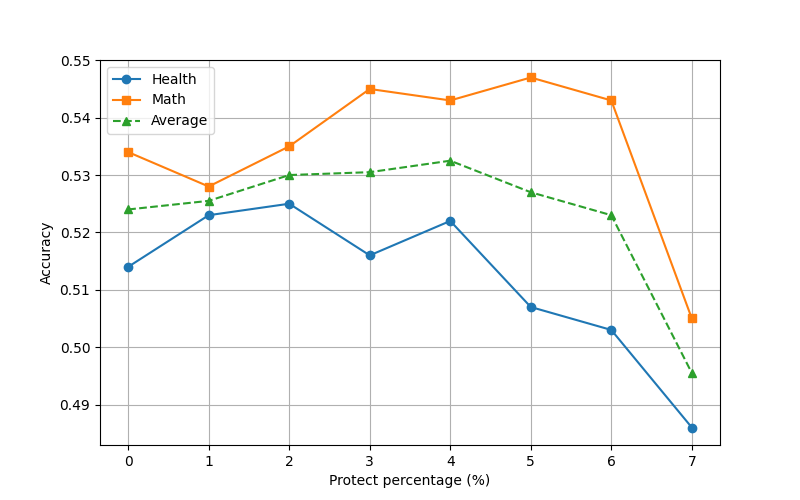}
\caption{Accuracy of average-merged models with FI-guided protection across both domains for different protection percentages $k$.}
\label{fig: protect}
\end{figure}

Figure~\ref{fig: protect} uses average merging as an example. The results indicate that as the percentage of protected parameters increases, the performance of the merged models initially improves but later declines, highlighting a trade-off in FI-guided protection. Protecting a small proportion of parameters with the highest FI helps mitigate performance degradation caused by parameter conflicts. However, a high percentage of protection may lead to forgetting issues in both domains. To determine the optimal protection percentage, we conduct a hyperparameter search on the validation set. More details can be found in Appendix~\ref{Appendix: merge}.

\section{Conclusion, Limitations \& Discussion}
 % Limitation requrired by ACL.
In summary, we introduced a stability measure, FI, to systematically identify the fragility of LLMs and VLMs (Breach in the shield). Through experiments under both internal and external perturbations, we demonstrate the effectiveness of our proposed method.

Our work constitutes an initial attempt to leverage sensitivity measures for improving model performance, focusing primarily on their application to model merging at the inference stage. While our study provides insights into the potential of such measures, we believe that further research is warranted to explore their utility in enhancing model training. 
%In particular, extending sensitivity-based metrics, such as FI, to guide parameter updates during fine-tuning or pre-training remains an open and promising direction for future work.

\section{Limitations}
Our method relies on gradient information and is not applicable to ``black-box'' models that do not expose internal parameters or gradients to users. In such cases, text-only approaches like Influence Function \citep{koh2017understanding} are more suitable.

\newpage
\bibliography{custom}

\appendix
\section{Appendices}
\subsection{Detail of Parameter sparsification experiment}
\label{Appendix: sparsification}

\textbf{Experiment on MMLU} We conduct experiments on the multiple-choice problems from the MMLU \citep{hendrycks2020measuring} dataset, using Qwen2-7B. We take the cross-entropy loss, \textit{i.e.}, $f = - \log P(y = y_{\text{pred}} | x, \theta)$, as the target function, and calculate the FI value according to Theorem ~\ref{FI-close}. In this setup, we treat the task as a 4-class classification problem with the possible classes being "A," "B," "C," and "D".

\textbf{Experiment on Alpaca-Eval} We use the Alpaca-eval validation set \citep{dubois2024length}, a widely adopted benchmark, and conduct experiments with various open-source models, including LLaMA2, LLaMA3 \citep{touvron2023llama}, and Qwen2 \citep{bai2023qwen}, across different sizes. We report two metrics: ROUGE-1 (comparing to pre-sparsity responses) and length-control winning rate (LCWR), comparing to GPT-3.5 Turbo. Higher scores are better for both metrics.

To estimate the average FI for sequence generation, we use the fixed-context approach with \( L = 5 \). For each sample \( z \), we estimate \( \mathbf{FI}_l(z) \) by generating \( N = 10 \) responses, truncating them at position \( l-1 \). These truncated sequences are used to approximate the conditional expectation by computing the sample average. The per-token FI values are then aggregated using Equation \ref{FI-seq} to obtain \( \mathbf{FI}^L_{\text{seq}}(z) \), which is averaged across all samples to estimate the overall FI.

\subsection{Computation complexity analysis}
\label{Appendix: computation}
    Let $n$ denote the number of samples, $p$ denote the dimension of perturbation ($p=3$ for pixel-wise computations and $p=1$ for parameter-wise computations), and $d$ represent the total number of pixels or parameters.
    
    \textbf{Computational Complexity Analysis:}
    \begin{itemize}
    \item \textbf{Jacobian-Norm}: $\mathcal{O}(npd)$, arising from gradient computation per pixel/parameter.
    \item \textbf{Saliency-Map}: Identical to Jacobian-Norm, $\mathcal{O}(npd)$.
    \item \textbf{FI-inverse}: $\mathcal{O}(np^3d + npd)$, with $\mathcal{O}(p^3)$ from inverse matrix computations and $\mathcal{O}(npd)$ from gradient calculations.
    \item \textbf{FI-cSVD (our method)}: $\mathcal{O}(np^2 r_0 d + npd)$, where $\mathcal{O}(p^2 r_0)$ stems from the compact SVD used to compute matrix inversion efficiently.
    \end{itemize}
    
    In practical scenarios:
    \begin{enumerate}
    \item \textbf{Parameter-wise stability}: Since individual parameters have dimension $p=1$, the FI calculation reduces to scalar inversion, thus the complexity simplifies to $\mathcal{O}(npd)$, matching Jacobian-Norm and Saliency-Map.
    \item \textbf{Pixel-wise stability (image data)}: Given that each pixel has dimension $p=3$ (RGB), the FI calculation involves compact SVD for a $3\times3$ matrix. Theoretically, this makes our method about $9$ times slower compared to baseline methods. However, in practical implementation, our approach is only approximately $2$ times slower.
    \end{enumerate}
    
    The table below presents the average time required to compute FI, Saliency Map, and Jacobian Norm for a single image using Qwen2VL-7B. All results are averaged over 100 images and measured on an A100-80G GPU.

    \begin{table}[h!]
    \centering
    \begin{tabular}{lc}
    \hline
    Method & Time (s) \\
    \hline
    FI     & 0.3828 \\
    Saliency-Map     & 0.1964 \\
    Jacobian-Norm    & 0.1939 \\
    \hline
    \end{tabular}
    \caption{Empirical computation times for different methods.}
    \label{tab:timing}
    \end{table}

\subsection{Reparametrization Invariance of FI}
\label{appendix: invariance}
The proposed FI measure has the property of transformation invariance.

\begin{theorem}[Reparametrization invariance]
    Suppose that \( \phi \) is a diffeomorphism of \( \omega \). Then, \( FI_{\omega}(\omega_0) \) is invariant with respect to any reparameterization corresponding to \( \phi \). Specifically, let
    \[
    \Tilde{\omega}(t) = \phi \circ \omega(t), \quad \Tilde{\omega}_0 = \phi(\omega_0),
    \]
    we have
    \[
    FI_{\Tilde{\omega}}(\Tilde{\omega}_0) = FI_{\omega}(\omega_0).
    \]
    \label{invariant}
    The detailed proof can be found in \cite{shu2019sensitivity}.
\end{theorem}

Theorem \ref{invariant} establishes that \( FI_{\omega}(\omega_0) \) is invariant under any diffeomorphic (e.g., scaling and spinning) reparameterization of the original perturbation. This invariance property is not shared by other measures, such as Jacobian norm \citep{novak2018sensitivity}, Cook's local influence measure \citep{cook1986assessment}, and Sharpness \citep{novak2018sensitivity}.  

For instance, consider a perturbation of the form \( \alpha + \Delta \alpha \), where \( \alpha \) is a subvector of \( (x^\top, \theta^\top)^\top \). If we apply a scaling reparameterization \( \alpha' = K \odot \alpha \), where \( K \) is a scaling vector and \( \odot \) denotes element-wise multiplication, then the Jacobian norms change:
\[
\|J(\alpha)\|_{F} = \left[\sum_{i} \left(\frac{\partial f}{\partial \alpha_i}\right)^{2} \right]^{1/2} \neq \|J(\alpha')\|_{F}.
\]
In contrast, the FI measure remains unchanged. Such a reparameterization does not alter the function itself but may affect the measure values, potentially weakening the correlation between perturbation and performance degradation. A similar discussion can be found in \citep{dinh2017sharp}.

\onecolumn

\subsection{Detail of FI-guided protection in model merging}
\label{Appendix: merge}

\begin{table}[h]
\centering
\caption{Searched ranges of hyperparameters of model merging methods}
\resizebox{\columnwidth}{!}
{
\begin{tabular}{|l|l|}
\hline
\textbf{Hyper parameter} & \textbf{Search Ranges of Hyperparameters} \\ \hline
Protecting ratio $k$ & [1\%, 2\%, 3\%, 4\%, 5\%, 6\%, 7\%, 8\%, 9\%, 10\%] \\ \hline
Weight parameter $\gamma$ in Task Arithmetic \& TIES& [0.3, 0.4, 0.5, 0.6, 0.9, 1.0] \\  \hline
\end{tabular}
}
\label{table: hyper}
\end{table}

\subsection{Proof of Theorem \ref{FI-close}}

\begin{proof} \label{proof2.4}
We apply Taylor expansion to $f({\omega}(t)) $ at the point ${\omega}(t)$: 
\begin{align*}
f({\omega}(t))&=f({\omega}(0))+{\nabla}_{f\left({\omega}_{0}\right)}^{T} {h}_{{\omega}_{0}} t + \frac{1}{2}\left({h}_{{\omega}_{0}}^{T} {H}_{f\left({\omega}_{0}\right)} {h}_{{\omega}_{0}}+{\nabla}_{f\left({\omega}_{0}\right)}^{T} d^{2} {\omega}(0) / d t^{2}\right) t^{2} 
+o\left(t^{2}\right),
\end{align*}
where ${\nabla}_{f\left({\omega}_{0}\right)}=\partial f({\omega}) /\left.\partial {\omega}\right|_{{\omega}={\omega}_{0}}$ and ${H}_{f\left({\omega}_{0}\right)} = \partial^{2} f({\omega}) /\left.\partial {\omega} \partial {\omega}^{T}\right|_{{\omega}={\omega}_{0}}$. From the definition of $S_C$, \( S_{C}^{2}(\omega_t, \omega_0) \) can be approximated as $S_{C}^{2}({\omega}_{t}, {\omega}_{0}) = t^{2} {h}_{{\omega}_{0}}^{T} G_{\omega_{0}} {h}_{{\omega}_{0}}+o\left(t^{2}\right).$ Based on l’Hˆopital’s rule, the stability measure FI from Equation\ref{fi1} can be rewritten as:
\begin{equation*}
\mathbf{FI}_{{\omega}}\left({\omega}_{0}\right) =\max _{{h}_{{\omega}}} \frac{{h}_{{\omega}}^{T} {\nabla}_{f\left({\omega}_{0}\right)} {\nabla}_{f\left({\omega}_{0}\right)}^{T} {h}_{{\omega}}}{{h}_{{\omega}}^{T} G_{\omega_{0}} {h}_{{\omega}}} .
\end{equation*}
We then reparameterize \( \omega \) to \( \tilde{\omega} = G_{\omega_0}^{-1/2} \omega \). According to Theorem \ref{invariant}, the stability measure \( \mathbf{FI} \) remains invariant under this reparameterization
\begin{equation*}
    FI_{\omega}(\omega_0) = FI_{\Tilde{\omega}}(\Tilde{\omega}_0) =\argmax_{h_{\Tilde{\omega}}}\frac{h_{\Tilde{\omega}}^\top G^{-1/2}_{\omega_0} \nabla_{f(\omega_0)}\nabla^\top_{f(\omega_0)}G^{-1/2}_{\omega_0}h_{\Tilde{\omega}}}{h_{\Tilde{\omega}}^\top h_{\Tilde{\omega}}}.
\end{equation*}
The maximization problem is now in the form of a Rayleigh quotient, which attains its maximum when \( h_{\tilde{\omega}} \) is proportional to \( G_{\omega_0}^{-1/2} \nabla_{f(\omega_0)} \). Substituting back into the Rayleigh quotient, we find:
\begin{align*}
\mathbf{FI}_{\omega}(\omega_0) &= \frac{ \left( G_{\omega_0}^{-1/2} \nabla_{f(\omega_0)} \right)^{T} G_{\omega_0}^{-1/2} \nabla_{f(\omega_0)} \nabla_{f(\omega_0)}^{T} G_{\omega_0}^{-1/2} \left( G_{\omega_0}^{-1/2} \nabla_{f(\omega_0)} \right) }{ \left( G_{\omega_0}^{-1/2} \nabla_{f(\omega_0)} \right)^{T} \left( G_{\omega_0}^{-1/2} \nabla_{f(\omega_0)} \right) } \\
&= \frac{ \nabla_{f(\omega_0)}^{T} G_{\omega_0}^{-1} \nabla_{f(\omega_0)} \nabla_{f(\omega_0)}^{T} G_{\omega_0}^{-1} \nabla_{f(\omega_0)} }{ \nabla_{f(\omega_0)}^{T} G_{\omega_0}^{-1} \nabla_{f(\omega_0)} } \\
&= \nabla_{f(\omega_0)}^{T} G_{\omega_0}^{-1} \nabla_{f(\omega_0)}.
\end{align*}

This concludes the proof.
\end{proof}

\subsection{Additional experiment results on parameter sparsification}
\label{Appendix: sparsify}
\begin{table*}[t!]
    \centering
    \caption{Performance of Different Models Based on Criteria with Full Value and Sparsity Percentages}
    \label{sparsity-seq}
    \resizebox{\textwidth}{!}{
    \begin{tabular}{llcccccccccc}
        \toprule
        \multirow{2}{*}{\textbf{Model}} & \multirow{2}{*}{\textbf{Criteria}} & \multirow{2}{*}{\textbf{Full}} & \multicolumn{2}{c}{\textbf{6\% Sparsity}} & \multicolumn{2}{c}{\textbf{8\% Sparsity}} & \multicolumn{2}{c}{\textbf{10\% Sparsity}} & \multicolumn{2}{c}{\textbf{12\% Sparsity}} \\
        \cmidrule(lr){4-5} \cmidrule(lr){6-7} \cmidrule(lr){8-9} \cmidrule(lr){10-11}
     &                                 & & \textbf{FI-High} & \textbf{Random} & \textbf{FI-High} & \textbf{Random} & \textbf{FI-High} & \textbf{Random} & \textbf{FI-High} & \textbf{Random} \\
        \midrule
\multirow{2}{*}{Llama2-13B} & Rouge-1 & 1.0 & 0.52 & $0.59 \pm 0.02$ & 0.4 & $0.43 \pm 0.06$ & 0.18 & $0.68 \pm 0.01$ & \underline{0.05} & $0.19 \pm 0.03$ \\
& LCWR & 0.43 & 0.38 & $0.41 \pm 0.03$ & 0.29 & $0.34 \pm 0.07$ & \underline{0.09} & $0.42 \pm 0.0$ & \underline{0.01} & $0.08 \pm 0.05$ \\
\midrule
\multirow{2}{*}{Llama3-8B} & Rouge-1 & 1.0 & 0.46 & $0.52 \pm 0.04$ & 0.21 & $0.41 \pm 0.06$ & \underline{0.09} & $0.25 \pm 0.04$ & \underline{0.04} & $0.12 \pm 0.03$ \\
& LCWR & 0.42 & 0.4 & $0.38 \pm 0.01$ & 0.12 & $0.30 \pm 0.03$ & \underline{0.0} & $0.12 \pm 0.01$ & \underline{0.0} & $0.01 \pm 0.01$ \\
\midrule
\multirow{2}{*}{Llama2-7B} & Rouge-1 & 1.0 & 0.44 & $0.56 \pm 0.01$ & 0.25 & $0.45 \pm 0.02$ & \underline{0.06} & $0.33 \pm 0.02$ & \underline{0.0} & $0.21 \pm 0.02$ \\
& LCWR & 0.42 & 0.32 & $0.4 \pm 0.0$ & 0.12 & $0.35 \pm 0.01$ & \underline{0.0} & $0.19 \pm 0.05$ & \underline{0.0} & $0.1 \pm 0.03$ \\
\midrule
\multirow{2}{*}{Qwen2-7B} & Rouge-1 & 1.0 & \underline{0.09} & $0.41 \pm 0.05$ & \underline{0.01} & $0.30 \pm 0.09$ & \underline{0.01} & $0.31 \pm 0.06$ & \underline{0.01} & $0.15 \pm 0.02$ \\
& LCWR & 0.41 & \underline{0.03} & $0.35 \pm 0.03$ & \underline{0.02} & $0.25 \pm 0.1$ & \underline{0.03} & $0.20 \pm 0.05$ & \underline{0.03} & $0.08 \pm 0.02$ \\
\midrule
\multirow{2}{*}{Qwen2-1.5B} & Rouge-1 & 1.0 & 0.18 & $0.4 \pm 0.13$ & 0.16 & $0.32 \pm 0.02$ & \underline{0.05} & $0.28 \pm 0.08$ & \underline{0.05} & $0.23 \pm 0.02$ \\
& LCWR & 0.14 & \underline{0.03} & $0.07 \pm 0.04$ & \underline{0.04} & $0.02 \pm 0.02$ & \underline{0.0} & $0.04 \pm 0.0$ & \underline{0.0} & $0.02 \pm 0.02$ \\
\bottomrule
    \end{tabular}
    }
\end{table*}

\end{document}